\documentclass[11pt]{article}

\usepackage[preprint]{acl}

\usepackage{times}
\usepackage{latexsym}

\usepackage{microtype}

\usepackage{inconsolata}

\usepackage[utf8]{inputenc} %
\usepackage[T1]{fontenc}    %
\usepackage{hyperref}       %
\usepackage{url}            %
\usepackage{booktabs}       %
\usepackage{amsfonts}       %
\usepackage{nicefrac}       %
\usepackage{microtype}      %
\usepackage{xcolor}         %

\usepackage{graphicx}
\usepackage{verbatim}
\usepackage{xspace}
\usepackage{amsmath}
\usepackage{listings}
\usepackage{adjustbox}
\usepackage{wrapfig}
\usepackage{tcolorbox}

\usepackage{longtable}
\usepackage[labelfont=bf]{caption}

\renewcommand{\sectionautorefname}{\S\kern-2pt}

\definecolor{keywordcolor}{RGB}{0,119,170}
\definecolor{stringcolor}{RGB}{186,33,33}
\definecolor{commentcolor}{RGB}{24,128,24}
\definecolor{backgroundcolor}{RGB}{248,248,248}
\definecolor{numbercolor}{RGB}{124,124,124}
\lstdefinestyle{mystyle}{
    backgroundcolor=\color{backgroundcolor},   
    commentstyle=\color{commentcolor},
    keywordstyle=\color{keywordcolor},
    numberstyle=\tiny\color{numbercolor},
    stringstyle=\color{stringcolor},
    basicstyle=\ttfamily\footnotesize,
    breakatwhitespace=false,         
    breaklines=true,                 
    captionpos=b,                    
    keepspaces=true,                 
    numbers=none,                    
    numbersep=5pt,                  
    showspaces=false,                
    showstringspaces=false,
    showtabs=false,                  
    tabsize=2,
    escapechar=|
}
\newcommand{\mdtick}{\textasciigrave}

\newcommand{\deepseeklarge}{DeepSeek Coder V2\xspace}
\newcommand{\deepseeklargefull}{DeepSeek Coder V2 Instruct\xspace}
\newcommand{\deepseeksmall}{DeepSeek Coder V2 Lite\xspace}
\newcommand{\deepseeksmallfull}{DeepSeek Coder V2 Lite Instruct\xspace}
\newcommand{\qwenlarge}{Qwen 2.5 Coder 32B\xspace}
\newcommand{\qwenlargefull}{Qwen 2.5 Coder 32B Instruct\xspace}
\newcommand{\qwensmall}{Qwen 2.5 Coder 7B\xspace}
\newcommand{\qwensmallfull}{Qwen 2.5 Coder 7B Instruct\xspace}
\newcommand{\codestral}{Codestral 22B\xspace}
\newcommand{\codestralfull}{Codestral 22B v0.1\xspace}
\newcommand{\llamasmall}{Llama 3.1 8B\xspace}
\newcommand{\llamasmallfull}{Llama 3.1 8B Instruct\xspace}
\newcommand{\llamalarge}{Llama 3.3 70B\xspace}
\newcommand{\llamalargefull}{Llama 3.3 70B Instruct\xspace}
\newcommand{\gemmasmall}{Gemma 3 12B\xspace}
\newcommand{\gemmasmallfull}{Gemma 3 12B It\xspace}
\newcommand{\gemmalarge}{Gemma 3 27B\xspace}
\newcommand{\gemmalargefull}{Gemma 3 27B It\xspace}
\newcommand{\gpt}{GPT-4.1\xspace}
\newcommand{\gptfull}{GPT-4.1 (2025/04/14)\xspace}

\newcommand{\semtrace}{\texttt{SemTrace}\xspace}
\newcommand{\semtracejs}{\texttt{SemTrace-JS}\xspace}
\newcommand{\semtracephp}{\texttt{SemTrace-PHP}\xspace}
\newcommand{\cruxeval}{\texttt{CRUXEval}\xspace}
\newcommand{\cruxevalx}{\texttt{CRUXEval-X}\xspace}
\newcommand{\cruxevaljs}{\texttt{CRUXEval-JS}\xspace}
\newcommand{\cruxevalphp}{\texttt{CRUXEval-PHP}\xspace}
\newcommand{\cruxevali}{\texttt{CRUXEval-I}\xspace}
\newcommand{\cruxevalo}{\texttt{CRUXEval-O}\xspace}

\newcommand{\codesearchnet}{\texttt{CodeSearchNet}\xspace}
\newcommand{\codesearchnetpython}{\texttt{CodeSearchNet-Python}\xspace}
\newcommand{\codesearchnetjs}{\texttt{CodeSearchNet-JS}\xspace}
\newcommand{\codesearchnetphp}{\texttt{CodeSearchNet-PHP}\xspace}

\title{Sense and Sensitivity: Examining the Influence of Semantic Recall on Long Context Code Understanding}

\author{
  Adam Štorek
  \hspace{0.25em}
  Mukur Gupta
  \hspace{0.25em}
  Samira Hajizadeh
  \hspace{0.25em}
  Prashast Srivastava
  \hspace{0.25em}
  Suman Jana \\
  Columbia University \\
  \texttt{\{astorek, suman\}@cs.columbia.edu} \\
  \texttt{\{mukur.gupta, sh4635, ps3400\}@columbia.edu}
}

\begin{document}

\maketitle

\begin{abstract}
Large language models (LLMs) are increasingly deployed for understanding large codebases, but whether they understand operational semantics of long code context or rely on pattern matching shortcuts remains unclear. We distinguish between lexical recall (retrieving code verbatim) and semantic recall (understanding operational semantics). Evaluating 10 state-of-the-art LLMs, we find that while frontier models achieve near-perfect, position-independent lexical recall, semantic recall degrades severely when code is centrally positioned in long contexts. We introduce semantic recall sensitivity to measure whether tasks require understanding of code's operational semantics vs. permit pattern matching shortcuts. Through a novel counterfactual measurement method, we show that models rely heavily on pattern matching shortcuts to solve existing code understanding benchmarks. We propose a new task \semtrace, which achieves high semantic recall sensitivity through unpredictable operations; LLMs' accuracy exhibits severe positional effects, with median accuracy drops of 92.73\% versus \cruxeval's 53.36\% as the relevant code snippet approaches the middle of the input code context. Our findings suggest current evaluations substantially underestimate semantic recall failures in long context code understanding.\footnote{Our code is available at \url{https://github.com/adamstorek/long-context-code-understanding}.}
\end{abstract}

\section{Introduction}
Large Language Models (LLMs) are increasingly applied to industry coding tasks~\citep{google_2024_ai_code} that demand understanding of large codebases~\citep{jimenez2024swebench}. Recent advances~\citep{dao2022flashattention, peng2024yarn, su2024rope} enable these models to process extremely long inputs, up to millions of tokens~\citep{openai2025gpt41}. However, a fundamental question remains unanswered: when models solve code understanding tasks, are they processing the specific code provided in context, or applying memorized patterns from pretraining? This distinction becomes critical as LLMs are deployed in production environments where they must handle novel, project-specific code that cannot be solved through pattern matching alone. Pattern matching shortcuts may also result in LLMs missing subtle vulnerabilities~\citep{ding2025primevul}.

We introduce a key distinction between two capabilities for code understanding in long contexts: \emph{lexical recall}, meaning the ability to locate and reproduce code verbatim, and \emph{semantic recall}, meaning the ability to remember what code does when it is run, i.e., its operational semantics~\citep{winskel1993formalsemantics}. These capabilities are distinct; models can have perfect lexical recall yet fail at semantic recall, demonstrating they can access relevant code but not understand its effects. While needle-in-the-haystack (NIAH) benchmarks~\citep{liu2024repoqa, liu2024repobench} measure lexical recall, the relationship between lexical and semantic recall is not well understood.

Moreover, code understanding tasks like output prediction are designed to measure semantic recall, but can often be solved through pattern-matching shortcuts (recognizing familiar algorithms, applying memorized correlations) without requiring semantic recall of the specific implementation. This conflation poses a fundamental evaluation challenge: existing benchmarks may allow shortcuts that mask semantic recall failures. We introduce \emph{semantic recall sensitivity} as a property of tasks: the degree to which solving the task requires semantically recalling specific code details rather than pattern matching.

To investigate whether lexical and semantic recall rely on different mechanisms, we leverage positional variation in long contexts as a diagnostic lens for how models integrate information, rather than as an end in itself. Positional variation naturally arises as code length scales, and understanding of code should remain stable regardless of where relevant details occur. Prior work has shown that LLM performance varies based on where information appears in the input~\citep{liu2023lostmiddlelanguagemodels, lu-etal-2022-fantastically}, but here we treat position as a controlled perturbation to probe representational differences between lexical and semantic recall. Our evaluation across 10 state-of-the-art LLMs reveals a clear dissociation: frontier models achieve near-perfect, position-independent lexical recall of relevant code, yet semantic recall on the same code degrades when centrally positioned within the code context.

On \cruxeval~\citep{gu2024cruxeval}, a popular input-output prediction benchmark, semantic recall shows moderate position-dependent degradation (median accuracy drop of 53.36\%) compared to negligible lexical recall degradation (2.39\%). However, we hypothesize this substantially underestimates semantic recall fragility due to low semantic recall sensitivity, as \cruxeval permits pattern matching that compensates for semantic recall failures. To test this, we propose a novel counterfactual measurement method: systematically removing lines from code and measuring the LLMs' performance degradation. Code that requires semantic recall should cause the LLMs' performance to fall sharply (like the Python interpreter), since the code's operational semantics eventually changes; LLMs using pattern matching degrade gradually. \cruxeval shows gradual degradation (only 44.15-59.74\% accuracy loss at 50\% line removal), confirming low sensitivity.

To isolate semantic recall, we introduce \semtrace, an output prediction task that achieves high semantic recall sensitivity through unpredictable operations. \semtrace reveals dramatically more severe degradation: median accuracy drops of 92.73\% versus \cruxeval's 53.36\%, with some frontier models reaching zero accuracy when code is centrally positioned. Even \gpt shows clear position-dependence when scaled to higher-digit arithmetic that cannot be memorized, confirming that moderate degradation on existing benchmarks masks severe semantic recall fragility.

Contributions. (1) We introduce the distinction between lexical recall (accessing code verbatim) and semantic recall (understanding operational semantics), demonstrating they dissociate in long contexts: models maintain near-perfect, position-independent lexical recall while semantic recall exhibits severe position-dependent failures. We validate this across multiple programming languages (Python, JavaScript, PHP). (2) We propose semantic recall sensitivity as a task property with a counterfactual measurement method, revealing existing benchmarks like \cruxeval have low sensitivity, allowing pattern matching to mask failures. (3) We introduce \semtrace, an output prediction task isolating semantic recall via unpredictable operations, revealing 92.73\% median accuracy degradation vs. 53.36\% on \cruxeval, demonstrating current evaluations substantially underestimate semantic recall challenges.

\section{Related Work}
Our work intersects a number of research areas, namely long context evaluation, positional effects in long contexts, and memorization.

\subsection{Long Context Evaluation}
Significant effort has been devoted to long-context evaluation, expanding from needle-in-haystack retrieval~\citep{mohtashami2023randomaccess, shaham-etal-2023-zeroscrolls, greg_kamradt_needle_2025} to comprehensive benchmarks~\citep{bai-etal-2024-longbench, bai2025longbenchv2deeperunderstanding}, extended context lengths~\citep{zhang-etal-2024-bench}, and complex reasoning tasks~\citep{kuratov2024babilong, hsieh2024ruler, levy-etal-2024-task, modarressi2025nolimalongcontextevaluationliteral}. For code specifically, prior work has focused on repository-level retrieval and code completion~\citep{guo2023longcoder, liu2024repobench}, semantic search~\citep{liu2024repoqa}, and code repository QA~\citep{bai2025longbenchv2deeperunderstanding}. However, these tasks primarily evaluate \emph{code retrieval} (locating relevant code given queries) rather than understanding \emph{operational semantics} (predicting execution behavior or input-output mappings). While prior work evaluates retrieval and understanding as separate tasks, we test both capabilities on identical code in long contexts to understand their relationship. We also introduce semantic recall sensitivity as a property for characterizing code understanding benchmarks.

\subsection{Position Effects in Long Context}
Prior work has identified positional biases in LLMs, ranging from long context scenarios~\citep{zhang2024found} to in-context learning~\cite{fang2025rethinking, min-etal-2022-rethinking, lu-etal-2022-fantastically}, with empirical and theoretical evidence suggesting the important role of positional encoding, loss functions, and data distribution~\citep{wu2025on, gu2025when}. Particularly relevant to our work is the 'lost-in-the-middle' effect~\citep{liu2023lostmiddlelanguagemodels, gao-etal-2024-insights}, where performance significantly degrades when relevant information is centrally positioned within the natural language input. However, prior work has not examined position effects on LLMs' understanding of operational semantics. We are the first to investigate how position affects code understanding, and, critically, to distinguish how it differentially impacts lexical vs. semantic recall.

\subsection{Memorization}
LLMs have been shown to have memorized numerous algorithms, facts, and patterns from pretraining~\citep{hartmann2023sokmemorizationgeneralpurposelarge, chang-etal-2023-speak, nanda2023progress}, which can mask their true reasoning ability and hinder evaluation~\citep{huang-chang-2023-towards}. To counter memorization, prior work perturbs 
problems such that their solutions change: 1-indexed Python~\cite{wu-etal-2024-reasoning}, mutated LeetCode problems~\cite{yang2025codereasoning}, or altered math problems~\citep{huang2025mathperturb}. We propose a complementary counterfactual approach: rather than perturbing code to alter results, we systematically remove information (one line at a time) from code snippets to ultimately render them unsolvable. If accuracy remains stable despite missing critical information, the model is relying on memorized patterns rather than the provided code.

\section{Lexical and Semantic Recall}
We distinguish between two types of recall capabilities that become critical when evaluating code understanding in long contexts. The first is~\emph{Lexical Code Recall ($R^L$)}, defined as the ability to reproduce code snippets from the input context verbatim, without necessarily understanding their meaning or behavior. This capability can be directly assessed through retrieval or cloze-style tasks, and modern LLMs achieve near-perfect performance on such benchmarks~\citep{qwen2025qwen25technicalreport, openai2025gpt41}.

The second is~\emph{Semantic Code Recall ($R^S$)}, defined as the ability to understand what a code snippet from the input context does when it is run: its operational semantics~\citep{winskel1993formalsemantics}. $R^S$ measures the~\emph{usability} of the model's representation for performing downstream reasoning tasks that depend on the provided code. For instance, given an in-context implementation of quicksort and an unsorted array, $R^L(quicksort)$ corresponds to regenerating the quicksort code verbatim, whereas $R^S(quicksort)$ corresponds to leveraging understanding of that code to correctly predict a sorted output array by tracing quicksort's execution steps.

These capabilities are distinct: models can have perfect lexical recall yet fail at semantic recall, demonstrating they can access relevant code but not process it. While needle-in-the-haystack benchmarks~\citep{liu2024repoqa, liu2024repobench} measure lexical recall, no prior work has systematically investigated semantic recall (understanding operational semantics through tasks like input-output prediction) in long contexts. Moreover, the relationship between these capabilities remains poorly understood.

\subsection{Semantic Recall Sensitivity}
We introduce~\emph{semantic recall sensitivity} as a property of a code understanding task quantifying the degree to which a task requires understanding operational semantics of a given code snippet rather than permitting pattern matching shortcuts.~\emph{Pattern matching shortcuts} refer to solving tasks by applying memorized algorithmic patterns or spurious input-output correlations from pretraining, thereby bypassing semantic recall of the specific provided code. The semantic recall sensitivity of a benchmark is the average sensitivity across its examples.

Consider an output prediction task where a model must predict what a function returns given some input. If the function implements a standard algorithm like quicksort, a model might predict the sorted output by recognizing the algorithm's structure and applying memorized sorting behavior, without semantically recalling the specific implementation details. Such a task exhibits~\emph{low semantic recall sensitivity}.
Conversely, if the function implements a novel or arbitrary computation (e.g., a specific non-standard permutation of the input array defined only within the context), memorized patterns provide no shortcuts; instead, the model must semantically recall the specific code lines.
This task would have~\emph{high semantic recall sensitivity}, as pre-trained knowledge alone is insufficient.

This distinction has critical implications for evaluation. Low semantic recall sensitivity is inherently problematic for two reasons. First, in production environments, models encounter novel, project-specific code where pattern matching shortcuts are unavailable. Benchmarks permitting shortcuts fail to assess this critical capability. Second, the difference between correct and vulnerable code often lies in a single line: \citet{ding2025primevul} show models struggle to distinguish correct from vulnerable implementations when differences are subtle, suggesting pattern matching may be dangerous when small implementation details matter.

Moreover, low sensitivity masks semantic recall failures in long contexts. If pattern matching compensates for position-dependent semantic recall degradation, benchmarks will systematically underestimate the severity of understanding failures. We investigate this empirically in~\autoref{subsec: lexical vs semantic recall}, demonstrating that low-sensitivity benchmarks exhibit smoother degradation curves that obscure severe underlying semantic recall fragility.

\subsection{Measuring Semantic Recall Sensitivity}
\begin{figure}
    \centering
    \includegraphics[width=0.9\linewidth]{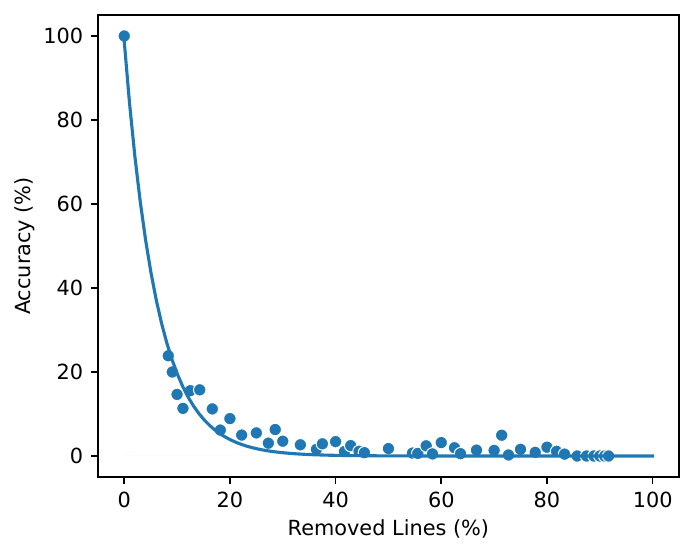}
    \caption{Python interpreter performance on incomplete \cruxeval functions. Accuracy drops sharply and approaches zero beyond 20\% removal, establishing the expected pattern when output prediction genuinely requires understanding all provided code. This provides a reference for evaluating LLM semantic recall sensitivity. Exponential trend fitted with bootstrapped 95\% CI.}
    \label{fig: incomp sensitivity python}
    \vspace{-1em}
\end{figure}
While semantic recall sensitivity characterizes a fundamental property of code understanding tasks, measuring it presents a challenge: how do we determine whether a model's success depends on understanding specific code versus applying pattern matching shortcuts?

We propose estimating semantic recall sensitivity by measuring the impact of withholding information from the code snippet. The key intuition is that for code reasoning tasks like output prediction, removing lines fundamentally changes the code such that it can no longer produce the correct output. Therefore, if a model continues to provide correct answers despite missing critical information, it cannot be reasoning about the specific provided code; instead, it must be relying on memorized patterns. Conversely, if performance degrades sharply as information is removed, the task requires genuine semantic recall of the specific code details.

Formally, let $Inc(C)$ denote a set of incomplete versions $C^{inc}$ from the original snippet $C$ (e.g., by removing specific lines). We define the sensitivity of task $\mathcal{T}$ with respect to $C$, denoted $Sens^{\mathcal{T}}(C)$, as the average normalized performance degradation across these incomplete versions (\autoref{eq:sensitivity_measure}):
\begin{equation}
\label{eq:sensitivity_measure}
    \frac{1}{|Inc(C)|} \sum_{C^{inc} \in Inc(C)} \frac{R^{\mathcal{T}}(C) - R^{\mathcal{T}}(C^{inc})}{R^{\mathcal{T}}(C) + \epsilon}
\end{equation}

Here, $R^{\mathcal{T}}(C)$ is the model's reasoning performance on the original snippet, and $R^{\mathcal{T}}(C^{inc})$ is its performance on an incomplete version. The term inside summation captures the relative performance caused by the incomplete information in $C^{inc} \in Inc(C)$ (with $\epsilon > 0$ added for numerical stability). Geometrically, $Sens^{\mathcal{T}}(C)$ measures the area under the performance degradation curve as information is progressively removed from $C$. A higher $Sens^{\mathcal{T}}(C)$ indicates a steeper degradation, meaning that successfully performing task $\mathcal{T}$ on snippet $C$ heavily depends on semantically recalling the specific details within $C$, thus signifying high semantic recall sensitivity. Operationally, our estimator is model-dependent; we aggregate it across models to characterize benchmark-level sensitivity.

To validate this approach and establish a reference point, we measure the Python interpreter's performance on incomplete CRUXEval functions. As expected, accuracy drops sharply--76 percentage points after removing just one line--and approaches zero beyond 20\% removal (\autoref{fig: incomp sensitivity python}). This exponential decay confirms that output prediction fundamentally requires semantic understanding of all code lines. Models showing gradual rather than sharp degradation therefore demonstrate low semantic recall sensitivity, indicating reliance on pattern matching rather than true code understanding.

\subsection{Measuring Semantic Recall: \semtrace}
\begin{figure}
\newcommand{\semtraceoutput}{$[38, 169, 16, 7]$\xspace}
\begin{lstlisting}[language=Python, caption={Example from \semtrace. Given input $x=81$, the model must predict the output \semtraceoutput by correctly recalling and applying each assignment line.}, label={fig:semtrace_example}]
def f(x):
    arr = [0, 0, 0, 0]
    arr[0] = x - 43
    arr[2] = x - 65
    arr[1] = x + 88
    arr[3] = x - 74
    return arr
\end{lstlisting}
\vspace{-0.25in}
\end{figure}

Having established how to measure semantic recall sensitivity, we now introduce a task designed to achieve high sensitivity by construction. To prevent pattern matching shortcuts, the task must require understanding implementation-specific details that cannot be inferred from algorithmic patterns or memorized correlations. We propose \semtrace, an output prediction task where errors can be directly attributed to semantic recall failures rather than general reasoning deficiencies. In \semtrace, the model receives a Python function that takes an integer $x$ as input, initializes a list, and populates it through assignment statements before returning the list. Crucially, each assignment modifies a distinct list element by adding a value $y$ (drawn uniformly from $[-100, 99]$) to the input $x$, and the order of assignment lines is randomized. An example is shown in~\autoref{fig:semtrace_example}.

We use simple 2-digit addition and subtraction operations ($x+y$) to minimize reasoning confounds while preventing pattern matching. These operations are simple enough that frontier LLMs can perform them reliably in isolation, ensuring that reasoning difficulty does not obscure semantic recall effects (we verify this empirically in~\autoref{tab: baseline perf}). However, because operations follow no predictable pattern, models cannot rely on pretrained heuristics without memorizing solutions for all possible combinations. The probability of guessing a single operation correctly is $\frac{1}{200}$; for functions with $k$ assignments ($k$ drawn uniformly from $[4, 10]$), the probability of guessing the entire output array by chance, even assuming knowledge of the generation process, is negligible (at most $(\frac{1}{200})^4 = 6.25 \times 10^{-10}$). Correctly predicting the output therefore necessitates accurate $R^S$ of all assignment lines, rendering \semtrace highly sensitive to $R^S$.

Since \semtrace's lines are mutually independent and comparable in difficulty, we can measure the degree of $R^S$ through partial matches: counting how many list positions the model predicted correctly. This metric allows us to distinguish semantic recall failures (some positions correct, indicating partial understanding) from complete breakdowns (near-random performance), confirming that errors reflect failures to recall specific lines rather than inability to perform the underlying operations.

\section{Experimental Setup \label{sec: experimental setup}}
\noindent{\textbf{Models.}}
We evaluate 10 state-of-the-art LLMs including \gptfull and five open-weight frontier models (\deepseeklargefull, \gemmalargefull, \llamalargefull, \qwenlargefull, and \codestralfull), and their smaller counterparts (\gemmasmallfull, \llamasmallfull, \qwensmallfull, \deepseeksmallfull) to examine scaling effects. Main results focus on the six frontier models for clarity; we include smaller model results in~\autoref{app: small model results}. Open-weight models were served on a \texttt{p5e.48xlarge} AWS EC2 instance using vLLM~\cite{kwon2023efficient}. Following~\citet{liu2023lostmiddlelanguagemodels}, we use greedy decoding for reproducibility. To manage costs, \gpt experiments use one-eighth of each dataset, randomly sampled.

\noindent{\textbf{Experimental Design.}} To isolate positional effects, we embed target code within contexts of irrelevant distractor code. Distractor functions are sampled from \codesearchnetpython~\citep{husain2019codesearchnet}, filtered by character count (25th–75th percentile, $\sim$200 tokens each) to exclude outliers. We use 20, 40, 60, or 80 distractors ($\sim$4k to $\sim$16k total tokens), systematically varying the target code's position across 11 equally spaced locations within each context. Crucially, distractors are unrelated to the target code, minimizing potential confounds from semantic interference while allowing us to cleanly measure position and context length effects.

\noindent{\textbf{Tasks.}}
We evaluate three capabilities: (1) semantic recall via \cruxeval~\citep{gu2024cruxeval} input and output prediction (800 Python functions each); (2) lexical recall via function-level retrieval on the same \cruxeval functions, where models must reproduce entire function bodies verbatim; (3) high-sensitivity semantic recall via \semtrace output prediction (800 generated functions).

For lexical recall, we prepend each line with a unique 6-digit hexadecimal key following key-value retrieval approaches~\citep{longchat2023, liu2023lostmiddlelanguagemodels}. This provides unambiguous reference markers and prevents models from identifying targets through naming patterns, which could confound position effects. For semantic recall sensitivity measurement, we generate incomplete versions of \cruxeval functions by systematically removing lines. We report zero-shot, exact match accuracy for all tasks. In~\autoref{fig: long context panel}, we report relative accuracy change, computed as percent change relative to each model's maximum performance at that context length, to facilitate comparison across models with different baseline capabilities.

We additionally verify our findings generalize across languages using four representative models. We draw from \cruxevalx \citep{xu2025cruxevalx}, an 18-language extension of \cruxeval, selecting JavaScript and PHP because \citet{xu2025cruxevalx} identify them as among the least correlated with Python, providing a conservative test of generalization. We apply the same experimental protocol with \cruxevaljs/\codesearchnetjs and \cruxevalphp/\codesearchnetphp. We also translate \semtrace to JavaScript and PHP following the same generation procedure, replacing Python-specific syntax with language-appropriate equivalents, yielding \semtracejs and \semtracephp.

\noindent{\textbf{Prompting.}}
Following \citet{liu2023lostmiddlelanguagemodels}, we use query-aware contextualization--placing the query both before and after the code--to enable decoder-only models to attend to the query when processing code. Prompt templates appear in~\autoref{app: prompt templates}.

\section{Results}
\autoref{tab: baseline perf} shows models' performance on \cruxeval and \semtrace without long-context distractors. Notably, frontier models achieve substantially higher median accuracy on \semtrace (86.19\%) than on \cruxeval input (45.06\%) or output prediction (55.38\%). Even \codestral and \gemmalarge, which perform comparatively worse on \semtrace, do not fail completely.
\begin{table}[h!]
\centering
\begin{adjustbox}{max width=\columnwidth}
\begin{tabular}{l||r|r}
\toprule
 \textbf{Model}                           & \textbf{\texttt{CRUXEval-I/O}} & \textbf{\semtrace}   \\
\midrule
 Codestral 22B                   & 44.62 / 50.38 &  23.25 \\
 \gpt                        & 79.00 / 77.00 & 100.00 \\ 
 DeepSeek Coder V2               & 36.50 / 54.00 &  96.00 \\
 Gemma 3 27B                     & 50.62 / 25.87 &  12.25 \\
 Llama 3.3 70B                   & 18.25 / 56.75 &  87.12 \\
 Qwen 2.5 Coder 32B               & 45.50 / 61.75 &  85.25 \\
 \bottomrule
\end{tabular}
\end{adjustbox}
\caption{Baseline performance (\% accuracy) without distractors. \texttt{CRUXEval-I/O} shows input/output prediction accuracy. Median accuracy is higher for \semtrace (86.19\%) than \cruxeval (45.06\% - 55.38\%), indicating its severe position-dependent degradation (\autoref{fig: long context panel} (c)) reflects high semantic recall sensitivity rather than inherent task difficulty.}
\label{tab: baseline perf}
\vspace{-1em}
\end{table}

\subsection{Lexical Recall Remains Stable While Semantic Recall Degrades with Position \label{subsec: lexical vs semantic recall}}
\begin{figure*}
    \centering
    \includegraphics[width=1.0\linewidth]{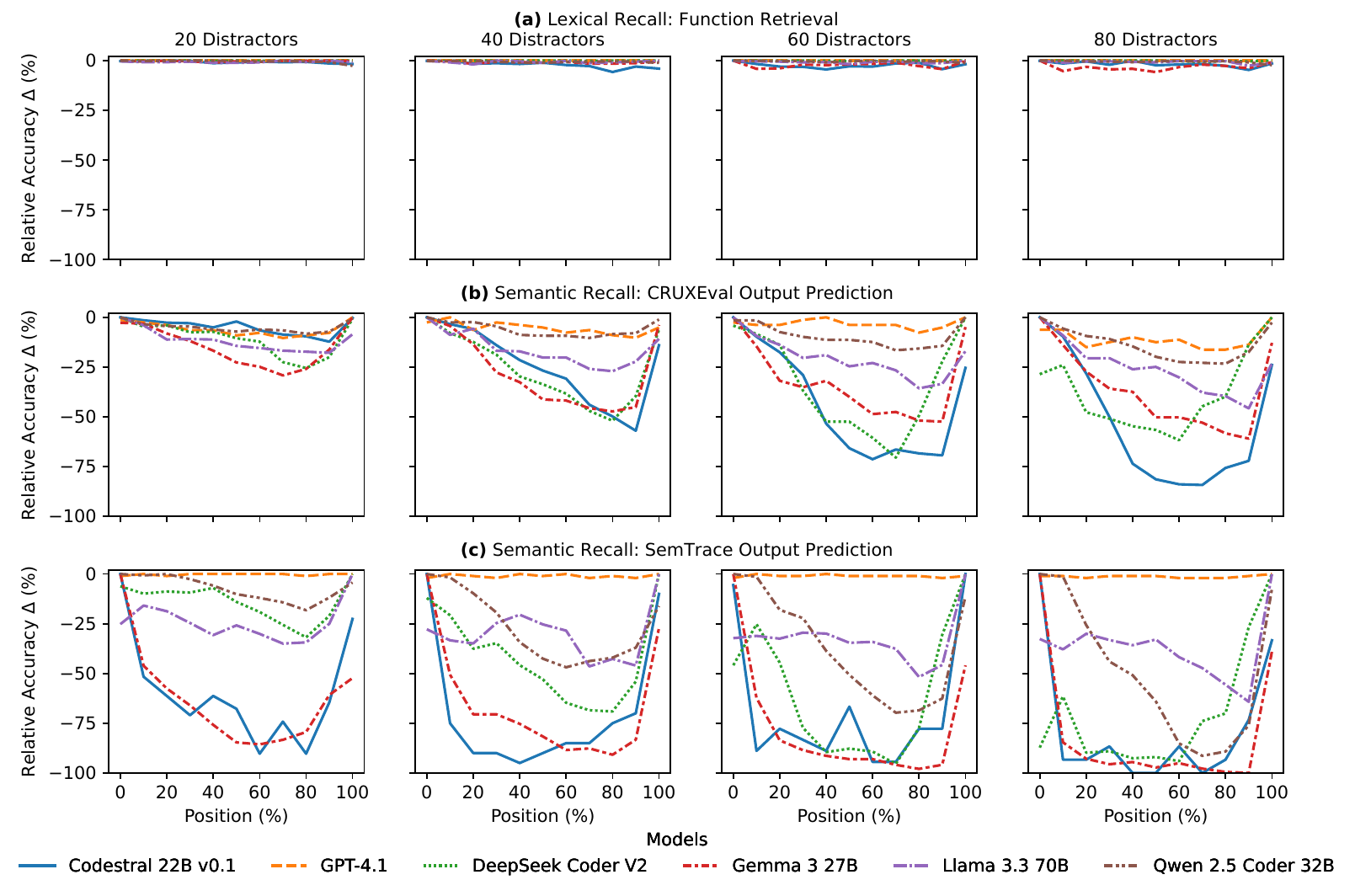}
    \caption{Dissociation between lexical and semantic recall across code positions in long contexts. Values show percent change relative to each model’s maximum accuracy at the same context length. (a) Lexical recall (function retrieval) remains near-perfect and position-independent across all context lengths for frontier models. (b) Semantic recall on \cruxeval output prediction (\cruxevalo) exhibits moderate lost-in-the-middle effects, with median accuracy decreases of 53.36\% as context length increases and target code moves toward the middle. (c) High-sensitivity semantic recall (\semtrace) shows severe position dependence, with median accuracy plummeting up to 92.73\% when target code is centrally positioned.} %
    \label{fig: long context panel}
\end{figure*}
\begin{figure*}
    \centering
    \includegraphics[width=1.0\linewidth]{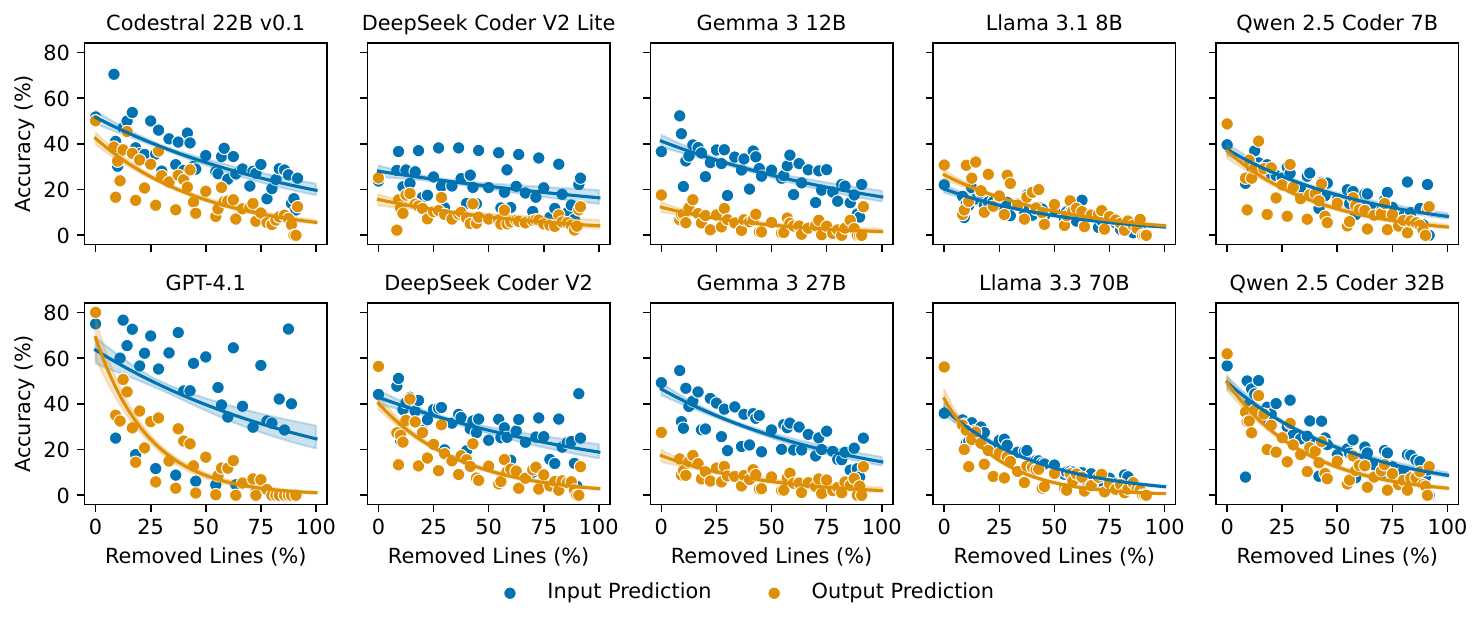}
    \caption{\cruxeval semantic recall sensitivity measurement across all models. Performance on input (blue) and output (orange) prediction tasks as lines are progressively removed from functions. We fit exponential trend lines using non-linear least squares and leverage bootstrapping to compute 95\% confidence intervals. Even with 50\% of lines removed, models' relative accuracy drops only 44.15\% and 59.74\% for input and output prediction, respectively, showing gradual degradation that contrasts sharply with the Python interpreter's exponential decay (\autoref{fig: incomp sensitivity python}). This gradual degradation indicates low semantic recall sensitivity, meaning models compensate for missing code through pattern matching from pretraining rather than relying on semantic understanding of the specific provided code.}
    \label{fig: incomp sensitivity}
\end{figure*}

\autoref{fig: long context panel} presents our central finding: a stark dissociation between lexical and semantic recall as code position varies in long contexts.

\textbf{Lexical recall is position-independent.} \autoref{fig: long context panel}a shows frontier models maintain over 95\% accuracy on function retrieval regardless of where target code appears, demonstrating they can reliably access code from any position within the tested context windows.

\textbf{Semantic recall degrades with position, moderately on \cruxeval.} \autoref{fig: long context panel}b reveals accuracy dropping as target code moves toward the center, with relative accuracy decreasing 16.25-84.29\% from maximum. Performance declines most severely when code appears at the 60-80\% position, while remaining relatively stable near context boundaries. This pattern is already present with only 20 distractor contexts ($\sim$4k tokens), becoming more pronounced as the code context size increases. \cruxeval input prediction exhibits a similar pattern (see~\autoref{app: cruxeval input pred results} for details).

\textbf{Semantic recall degrades severely on \semtrace.} \autoref{fig: long context panel}c shows dramatically intensified degradation compared to \cruxeval: \semtrace accuracy plummets when target code is centrally positioned. For example, \qwenlarge, the second-best performer on \cruxevalo with only 23.31\% relative degradation in accuracy for an 80-distractor code context ($\sim$16k tokens), suffers 91.38\% relative accuracy loss on \semtrace, dropping 53 percentage points; \codestral and \gemmalarge reach zero accuracy. This severe degradation appears even at shorter contexts and deepens substantially as context grows. Partial match analysis (\autoref{app: semtrace detailed analysis}) confirms this represents gradual semantic recall degradation rather than complete failure.

\textbf{Lexical-semantic recall dissociation generalizes across programming languages.} We consistently observe the same fundamental dissociation on JavaScript and PHP versions of \cruxeval and \semtrace: lexical recall remains generally stable across positions while semantic recall degrades as target code moves toward the center of the context, with median relative accuracy drops of 34.72\% and 14.35\% on \cruxevaljs and \cruxevalphp, and 80.39\% and 76.12\% on \semtracejs and \semtracephp, respectively, across four representative models (\autoref{app: crosslingual analysis}). On \semtracephp, several models achieve zero baseline accuracy despite performing better on \cruxevalphp than \cruxevaljs, suggesting the degree of pattern-matching reliance may vary across programming languages. These results signal that the phenomenon is not an artifact of Python-specific syntax or training.

\textbf{Out-of-distribution inputs cannot explain the difference in severity of positional degradation between \cruxeval and \semtrace}. If \semtrace introduced out-of-distribution inputs, models would struggle on \semtrace regardless of context length, and models that perform worse on \semtrace than on \cruxeval in short-context settings should diverge in positional behavior from those that perform better. We observe neither: most frontier models perform substantially better on \semtrace than on \cruxeval in short-context settings (\autoref{tab: baseline perf}), and, with the exception of \gpt, models with widely varying short-context \semtrace performance exhibit the same qualitative positional degradation pattern. We analyze \gpt separately in~\autoref{subsec: gpt exception}.

\textbf{Retrieval failures cannot explain semantic degradation}. Models demonstrably access code near-perfectly (\autoref{fig: long context panel}a) yet fail to understand its operational semantics (\autoref{fig: long context panel}b-c). We next explain why \cruxeval understates this degradation.

\subsection{\cruxeval's Low Semantic Recall Sensitivity Masks Position Effects}
Why does semantic recall show such different positional sensitivity between \cruxeval (moderate degradation) and \semtrace (severe degradation)? We hypothesize that \cruxeval has low semantic recall sensitivity, meaning models can partially solve output prediction tasks by applying pattern matching shortcuts from pretraining rather than semantically recalling the specific provided code.

To test this, we generate incomplete versions of each \cruxeval function by removing all possible combinations of lines while preserving function signatures, creating 71,994 incomplete functions. We measure accuracy aggregated by percentage of lines removed. If models truly require semantic recall of all provided code, we should observe sharp exponential decay similar to the Python interpreter (\autoref{fig: incomp sensitivity python}). Conversely, gradual degradation indicates models are compensating through pattern matching shortcuts instead.

\noindent{Counterfactual measurement reveals gradual degradation instead of sharp decline.} \autoref{fig: incomp sensitivity} shows the results across all 10 evaluated models. Removing 50\% of lines--making the code fundamentally incomplete and unexecutable--causes only 44.15\% accuracy drop for input prediction and 59.74\% for output prediction, contrasting sharply with the Python interpreter's exponential decay.

\textbf{Larger models show slightly higher sensitivity but remain resilient.} Larger models (\gpt, \deepseeklarge, \llamalarge, \qwenlarge) exhibit sharper performance declines than smaller models, suggesting their greater reliance on semantic recall. However, even these models degrade far more gradually than the Python interpreter, maintaining substantial accuracy despite fundamentally incomplete code.

\textbf{The low sensitivity to missing information reflects low semantic recall sensitivity.} Models leverage knowledge of common algorithmic patterns--sorting, string manipulation, basic arithmetic--to predict plausible outputs without fully understanding the specific implementation, masking semantic recall failures. This explains why \cruxeval shows only moderate position effects (\autoref{fig: long context panel}b) while \semtrace reveals severe degradation (\autoref{fig: long context panel}c).

\subsection{\gpt's Exception Reveals Memorization of Simple Arithmetic\label{subsec: gpt exception}}
\begin{figure}
    \centering
    \includegraphics[width=1.0\linewidth]{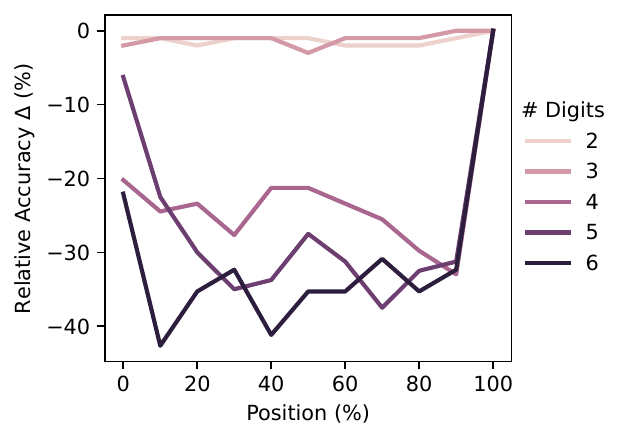}
    \caption{\gpt's higher-digit \semtrace performance reveals arithmetic memorization. Testing with 2--6 digit operations and 80 distractors shows position-independent performance for 2--3 digits, but clear position-dependent degradation for 4+ digits (a 31 percentage-point drop at 4 digits, 43\% relative loss at 6 digits). This demonstrates \gpt's perfect 2-digit performance (\autoref{fig: long context panel}c) reflects memorized arithmetic rather than superior semantic recall.}
    \label{fig: semtrace scaling}
    \vspace{-1em}
\end{figure}
\autoref{fig: long context panel} reveals an anomaly: \gpt achieves near-perfect accuracy on \semtrace regardless of position or context length, while still exhibiting clear position-dependent degradation on \cruxeval. This makes \gpt the only model where the pattern inverts, performing better on \semtrace than \cruxeval. If \gpt had truly superior semantic recall capabilities, why would it fail on the less sensitive task (\cruxeval) while succeeding on the more sensitive one (\semtrace)?

We hypothesize that \gpt has memorized 2-digit arithmetic, effectively converting \semtrace from a semantic recall task into a lexical recall task by simply retrieving memorized addition/subtraction results rather than reasoning about code. To test this, we evaluate \gpt on \semtrace variants using 3-, 4-, 5-, and 6-digit operations.

\autoref{fig: semtrace scaling} shows the results with 80 distractors ($\sim$16k tokens). \gpt maintains near-perfect performance on 2- and 3-digit \semtrace across all positions. However, position-dependent degradation emerges clearly for 4-digit operations, with accuracy dropping up to 31 percentage points when code is centrally positioned. This degradation intensifies for 5- and 6-digit operations, with 6-digit \semtrace showing up to 43\% relative accuracy loss and pronounced U-shaped curves.

This finding reveals that even high-sensitivity tasks remain vulnerable to shortcuts if they involve operations models can solve without processing code. \gpt's perfect 2-digit performance appeared to demonstrate superior semantic recall but exhibited position-independence characteristic of retrieval rather than processing: a pattern consistent with memorized arithmetic. The emergence of position-dependence at higher digits suggests that when such shortcuts become unavailable, even \gpt must rely on genuine semantic recall, which exhibits the same fragility as other frontier models.

\section{Broader Implications}
Although we study semantic recall in the context of code, the underlying capability of applying multi-step algorithms embedded in long contexts is not specific to programming. Many high-stakes NLP domains contain natural-language procedures with the same structural properties: a set of sequential, interdependent steps embedded in a large document that must be located, understood, and correctly applied to a specific case.

A concrete example is legal and policy analysis, where one must extract multi-step decision rules from large documents, such as ``if A, then check B and C; entitlement holds only if ...'', and apply them to specific cases. Indeed, \citet{jurayj2026taxreasoning} find that models struggle to apply relevant entries from a large corpus of U.S. federal tax statutory rules provided in-context, but perform substantially better when allowed to extract and encode the relevant rules into a symbolic solver that handles the procedural reasoning. This suggests that locating rules from long context and applying them are dissociable capabilities, consistent with the lexical-semantic distinction we document here.

More broadly, we view semantic recall sensitivity as a property relevant to any task where correct behavior requires understanding and applying specific procedural details from long contexts, rather than recognizing familiar patterns. We hope that the concepts, metrics, and benchmarks introduced here provide useful tools for investigating this phenomenon beyond code.

\section{Conclusion}
We introduced the distinction between lexical code recall (verbatim retrieval) and semantic code recall (understanding what code does), demonstrating that while frontier LLMs achieve near-perfect position-independent lexical recall, semantic recall exhibits severe lost-in-the-middle effects with median accuracy drops of 92.73\%. We proposed semantic recall sensitivity as a property of benchmarks and showed that existing code reasoning benchmarks like \cruxeval have low sensitivity compared to \semtrace, allowing pattern matching to mask the true severity of position-dependent failures. Our findings reveal that current long-context evaluations may substantially underestimate the challenges models face when reasoning about novel, unfamiliar code. Future work should develop benchmarks with high semantic recall sensitivity to more accurately assess models' true code understanding capabilities in long contexts.

\section*{Limitations}
\noindent{\textbf{Distractor Code Design.}} We use randomly sampled, semantically unrelated functions as distractors to isolate positional effects. While this design enables clean measurement of position-dependent degradation, real-world codebases contain semantically related functions that may introduce interference effects not captured in our study. While we anticipate this would lead to further deterioration, the relationship between semantic similarity of surrounding code and semantic recall failures remains an open question.

\noindent{\textbf{Task Scope.}} Our evaluation focuses on input-output prediction and code retrieval tasks. While these tasks directly measure semantic and lexical recall, production code understanding involves additional capabilities such as code generation, debugging, and complex multi-hop reasoning. The extent to which our findings apply to these broader tasks requires further investigation.

\noindent{\textbf{\semtrace Design Constraints.}} \semtrace uses simple arithmetic operations to isolate semantic recall from reasoning difficulty. While this design choice enables clear attribution of errors to semantic recall failures, it may not capture semantic understanding challenges in more complex algorithmic contexts. Additionally, our discovery that \gpt has memorized arithmetic operations highlights an inherent limitation: as models grow more capable, designing tasks that prevent all possible shortcuts becomes increasingly difficult.

\noindent{\textbf{Context Length Range.}} Our experiments evaluate contexts up to approximately 16k tokens. While this range covers common code understanding scenarios, recent models support significantly longer contexts (up to millions of tokens). While our results suggest increasing context length leads to increased semantic recall degradation, whether the pattern continues at extreme context lengths is left for future work.

\noindent{\textbf{Benchmark Scale.}} Our evaluations use 800 examples per task. While sufficient to observe consistent patterns across models, larger-scale evaluation could reveal more subtle effects and enable more fine-grained analysis of failure modes.

\section*{Acknowledgments}
We thank the anonymous reviewers as well as Andreas D. Kellas, Nihal Jain, and Abhishek Shah for their valuable feedback. This work was partially supported by an award from the Google Cyber NYC Institutional program. Any opinions, findings, conclusions, or recommendations expressed herein are those of the authors and do not reflect those of Google.

\bibliography{references} %

\appendix

\section{Results on Smaller Models \label{app: small model results}}
\begin{figure*}
    \centering
    \includegraphics[width=1.0\linewidth]{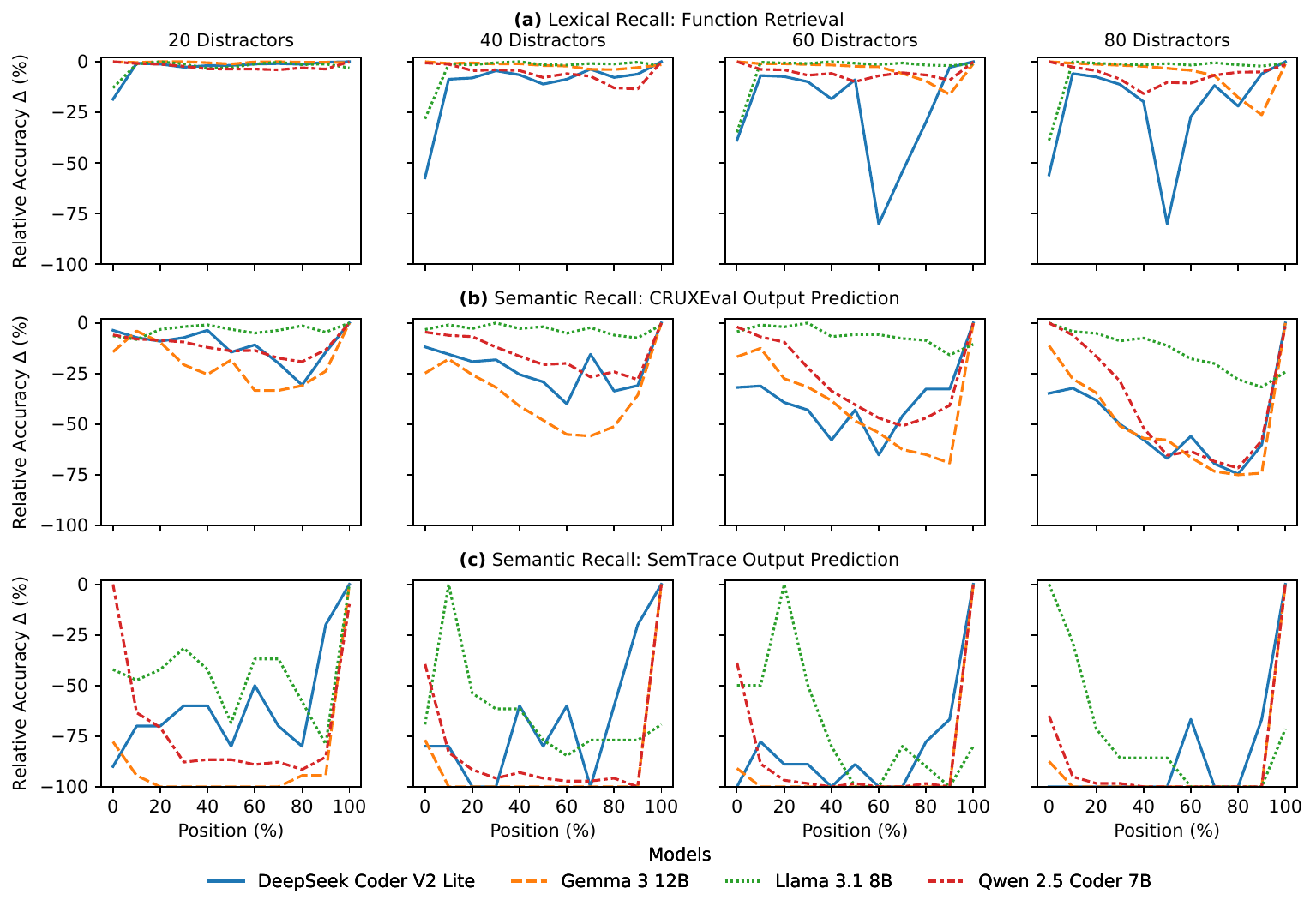}
    \caption{Dissociation between lexical and semantic recall across code positions in long contexts for smaller models. Positional effects across lexical recall (a), semantic recall on \cruxeval output prediction (\cruxevalo) (b), and high-sensitivity semantic recall on \semtrace (c). Despite lower absolute performance, smaller models exhibit the same fundamental dissociation as frontier models: position-independent lexical recall and position-dependent semantic recall, with more severe degradation on \semtrace.}
    \label{fig: long context panel small models}
\end{figure*}

\begin{table}[h!]
\small
\centering
\begin{tabular}{l||r|r}
\toprule
 \textbf{Model}                           & \textbf{\texttt{CRUXEval-I/O}} & \textbf{\semtrace}   \\
\midrule
 DeepSeek Coder V2 Lite          & 20.00/21.00 &   4.12 \\
 Gemma 3 12B                     & 39.50/19.25 &  15.88 \\
 Llama 3.1 8B                    & 19.50/32.62 &  16.00 \\
 Qwen 2.5 Coder 7B                & 31.62/46.62 &  22.25 \\
 \bottomrule
\end{tabular}
\caption{Baseline performance (\% accuracy) without distractors. \texttt{CRUXEval-I/O} shows input/output prediction accuracy. Smaller models generally struggle more on \semtrace relative to \cruxeval than their frontier counterparts, but similarly no model fails completely.}
\label{tab: baseline perf small models}
\end{table}
We evaluate four smaller models (\deepseeksmall, \gemmasmall, \llamasmall, and \qwensmall) across all tasks. While these models exhibit lower absolute performance than their frontier counterparts (\autoref{tab: baseline perf small models}), they demonstrate qualitatively similar patterns.~\autoref{fig: long context panel small models} shows that smaller models maintain the key dissociation between lexical and semantic recall: they achieve relatively stable lexical recall across positions while showing moderate and severe position-dependent semantic recall degradation on both \cruxeval and \semtrace, respectively.

However, smaller models exhibit notable differences in lexical recall stability compared to frontier models.~\autoref{fig: long context panel small models}a reveals greater positional variance and some anomalous accuracy drops, particularly \deepseeksmall showing degradation in the middle of the input context, and both \deepseeksmall and \llamasmall exhibiting drops at the beginning of the context. Despite these irregularities, the lexical recall patterns remain qualitatively distinct from the systematic lost-in-the-middle degradation observed in semantic recall tasks~\autoref{fig: long context panel small models}b-c).
Notably, the smaller models exhibit more pronounced performance variability and sharper degradation on \semtrace, suggesting their semantic recall capabilities are more fragile. However, the fundamental pattern (position-independent lexical recall vs. position-dependent semantic recall) remains consistent across model scales, confirming that our findings generalize beyond frontier models.

\section{\cruxeval Input Prediction (\cruxevali) Results \label{app: cruxeval input pred results}}
\begin{figure*}
    \centering
    \includegraphics[width=1.0\linewidth]{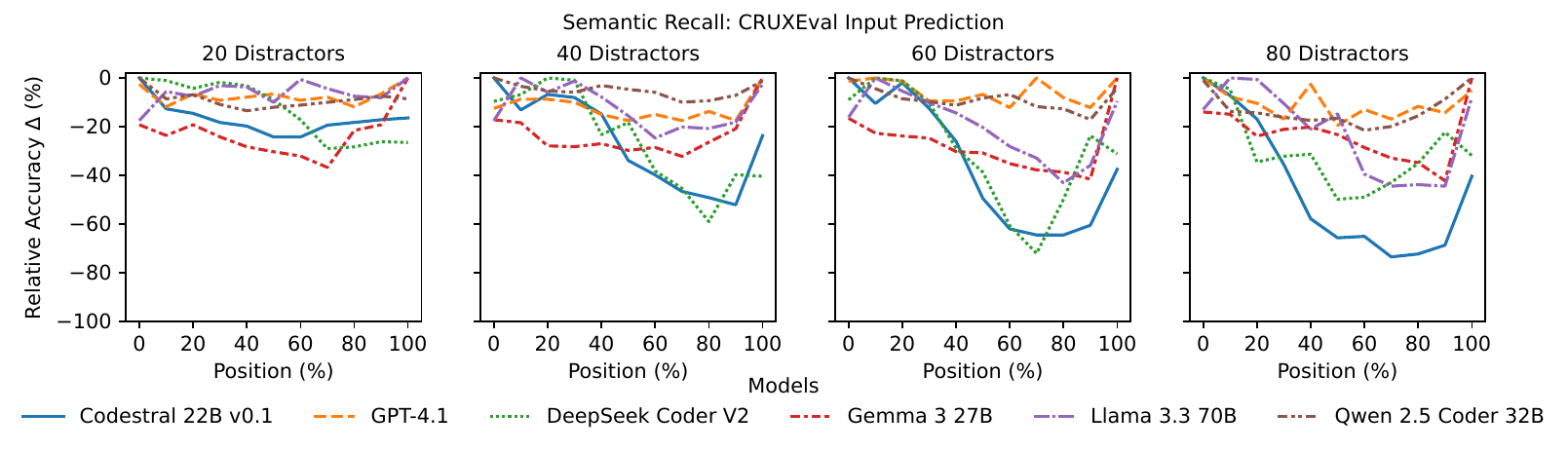}
    \caption{\cruxeval input prediction (\cruxevali) positional effects for frontier models. Similar to output prediction (\autoref{fig: long context panel}b), all models show moderate position-dependent degradation with the steepest declines occurring when target code appears at 60-80\% positions.}
    \label{fig: long context input prediction}
\end{figure*}
\begin{figure*}
    \centering
    \includegraphics[width=1.0\linewidth]{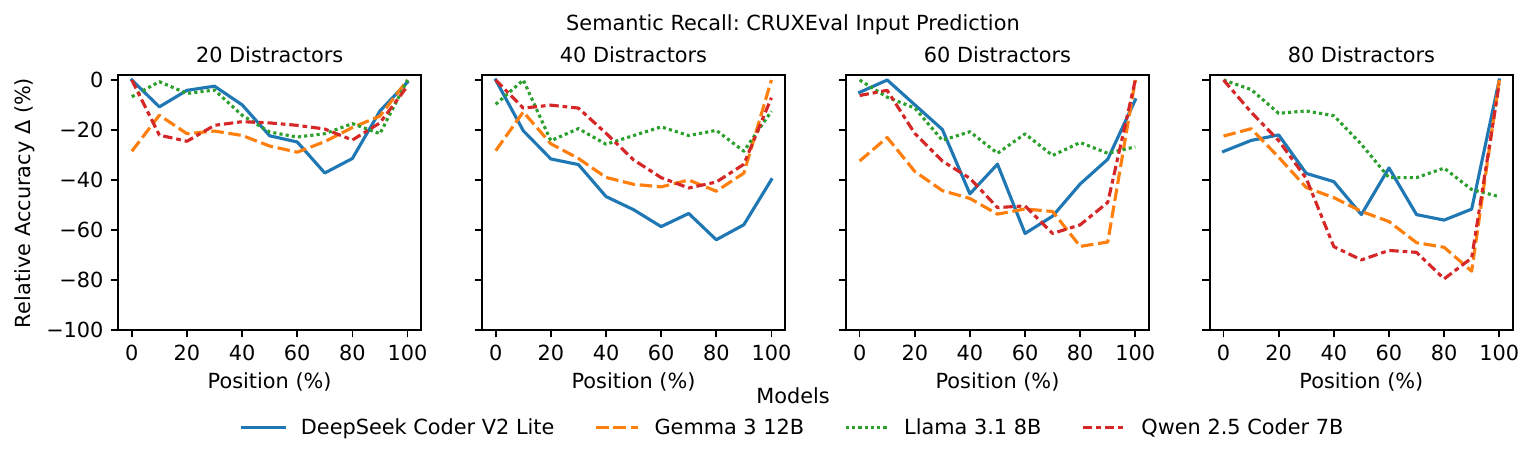}
    \caption{\cruxeval input prediction (\cruxevali) positional effects for smaller models. Performance patterns mirror those observed in \cruxeval output prediction (\autoref{fig: long context panel small models}b), with moderate lost-in-the-middle effects becoming more pronounced as context length increases.}
    \label{fig: long context input prediction small models}
\end{figure*}
We present complete results for \cruxeval input prediction (\cruxevali) to complement the output prediction (\cruxevalo) results shown in~\autoref{subsec: lexical vs semantic recall}.~\autoref{fig: long context input prediction} and~\autoref{fig: long context input prediction small models} demonstrate that models exhibit nearly identical positional sensitivity patterns on input prediction as they do on output prediction (\autoref{fig: long context panel}b and \autoref{fig: long context panel small models}b, respectively). Both tasks show moderate lost-in-the-middle effects, with performance degrading as target code moves toward central positions, particularly at the 60-80\% position range. This similarity is expected given that both tasks measure semantic recall on the same \cruxeval functions and are equally susceptible to pattern matching shortcuts, as confirmed by our counterfactual measurement (\autoref{fig: incomp sensitivity}).

\section{\semtrace Detailed Analysis \label{app: semtrace detailed analysis}}
\begin{figure*}
    \centering
    \includegraphics[width=1.0\linewidth]{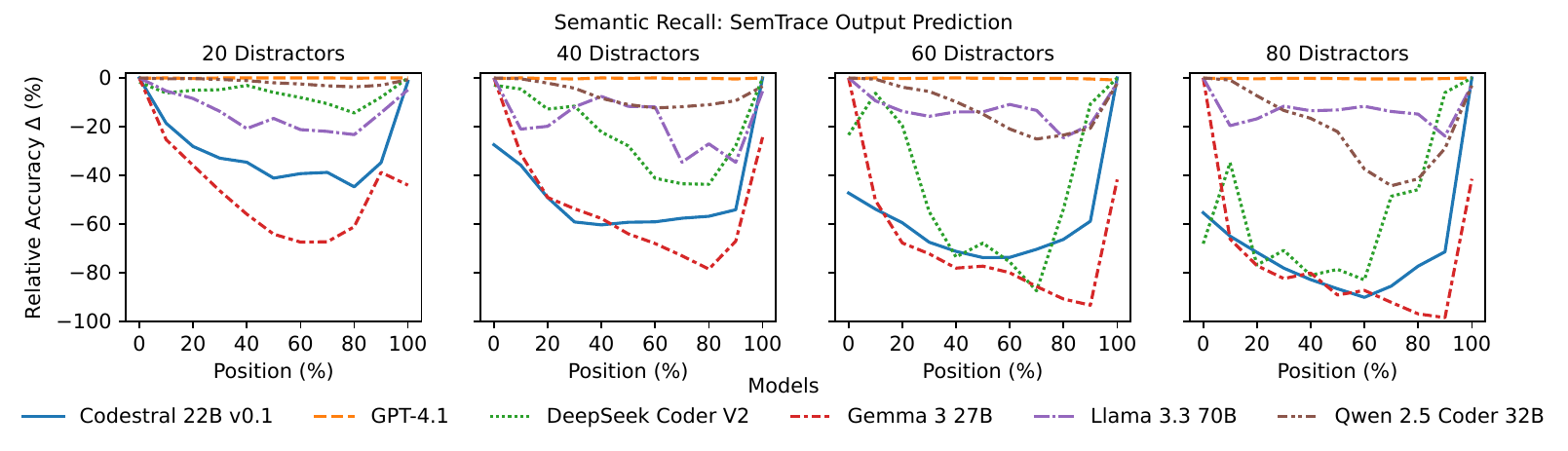}
    \caption{Partial match accuracy on \semtrace across code positions, measuring the percentage of list elements predicted correctly rather than requiring exact full-output matches. Despite following similar positional trends to exact match accuracy (\autoref{fig: long context panel}c), degradation curves are smoother, indicating that accuracy drops result from gradual semantic recall failures where models correctly process some assignment lines but not others, rather than complete task comprehension breakdown.}
    \label{fig: semtrace acc partial}
\end{figure*}

\noindent{\textbf{Partial Match Accuracy}} To further investigate whether the severe accuracy drops on \semtrace reflect complete task breakdown or gradual semantic recall failures, we measure partial match accuracy: counting how many individual list positions models predict correctly rather than requiring the entire output to be exact.~\autoref{fig: semtrace acc partial} reveals that partial match accuracy follows qualitatively similar positional patterns to exact match accuracy (\autoref{fig: long context panel}c), with performance degrading as target code moves toward central positions. However, the degradation curves are notably smoother, indicating that many errors in exact match accuracy stem from partial failures where models correctly recall some assignment lines but not others. This pattern confirms that position-dependent degradation reflects gradual semantic recall failures rather than complete comprehension breakdown, as models maintain the ability to semantically recall and apply some--but not all--assignment operations even under challenging positional conditions.

\begin{figure*}
    \centering
    \includegraphics[width=1.0\linewidth]{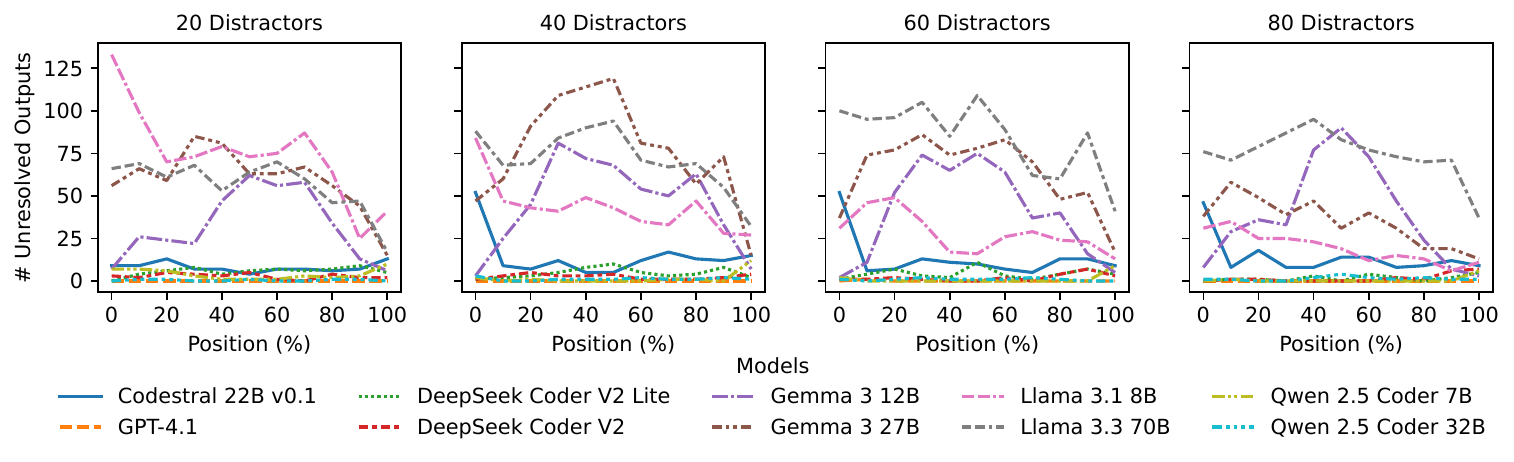}
    \caption{Number of unresolved outputs on \semtrace. Unresolved outputs are expressions like $81 - 43$ instead of the computed result $38$. Models from the Gemma and Llama families produce substantially more unresolved outputs when target functions appear near the middle of the context, suggesting an attempt to circumvent semantic recall limitations through lexical recall of the operations without performing the computation.}
    \label{fig: semtrace resolution error}
\end{figure*}

\noindent{\textbf{Resolution Errors}} Despite explicitly prohibiting partial results in the prompt, we find that several models still at times generate answers such as $81-43$ instead of $38$. Since these expressions are interpreted by Python as equivalent, we don't penalize such responses. However, our supplementary analysis (\autoref{fig: semtrace resolution error}) reveals that especially Gemma and Llama family models provide these "unresolved" outputs much more frequently when the target function appears near the middle of the input context. We hypothesize that this represents an attempt to circumvent semantic recall limitations through lexical recall.


\section{Extending to JavaScript and PHP \label{app: crosslingual analysis}}
\begin{figure*}
    \centering
    \includegraphics[width=1.0\linewidth]{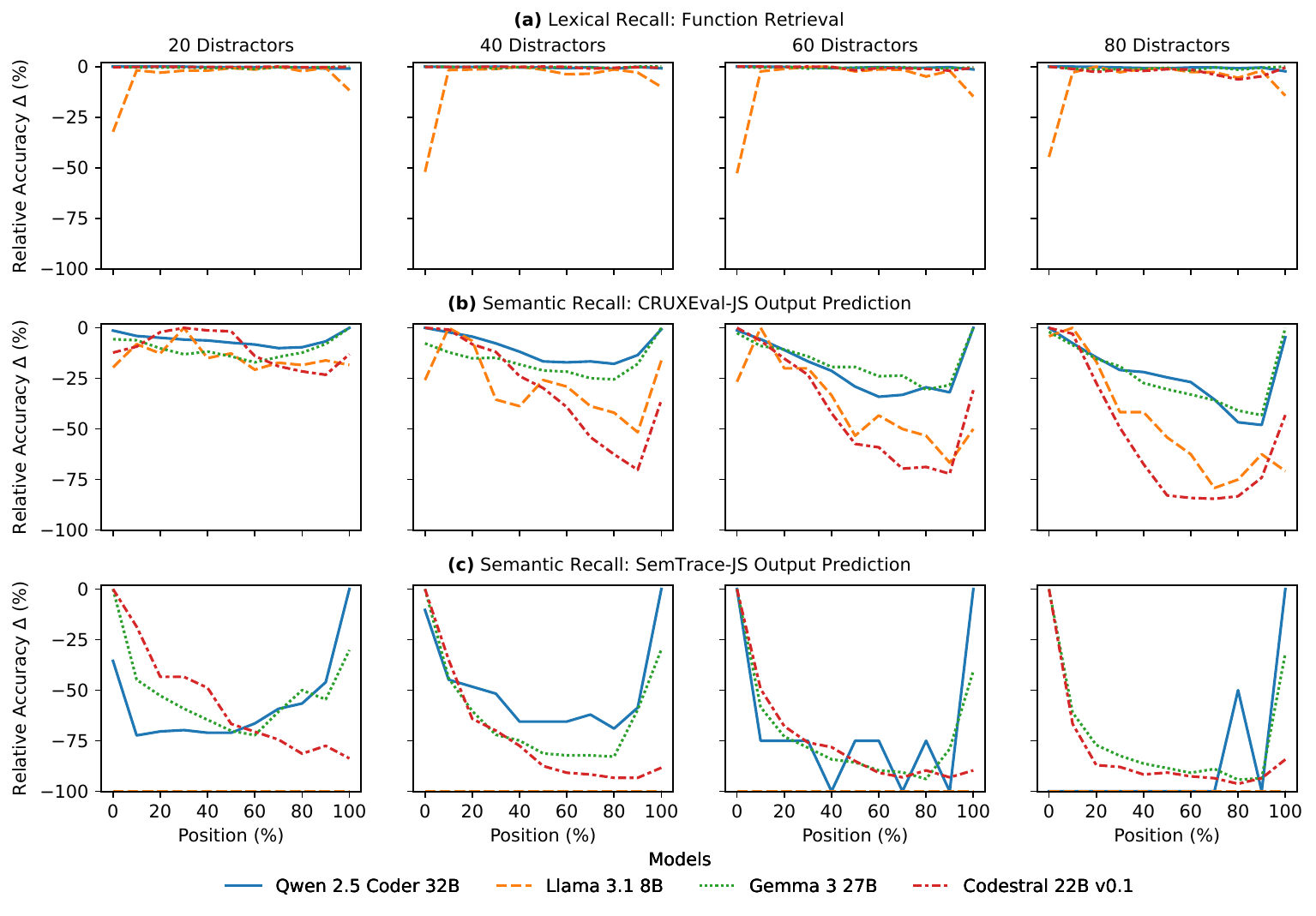}
    \caption{Dissociation between lexical and semantic recall across code positions in long contexts for JavaScript. Panel (a) shows lexical recall (function retrieval) remaining largely stable across positions, with the exception of \llamasmall which exhibits similar positional variance to its \cruxeval retrieval performance. Panels (b) and (c) show moderate and severe position-dependent semantic recall degradation on \cruxevaljs and \semtracejs, respectively, mirroring the pattern observed for Python in \autoref{fig: long context panel}.}
    \label{fig: long context panel js}
\end{figure*}

\begin{figure*}
    \centering
    \includegraphics[width=1.0\linewidth]{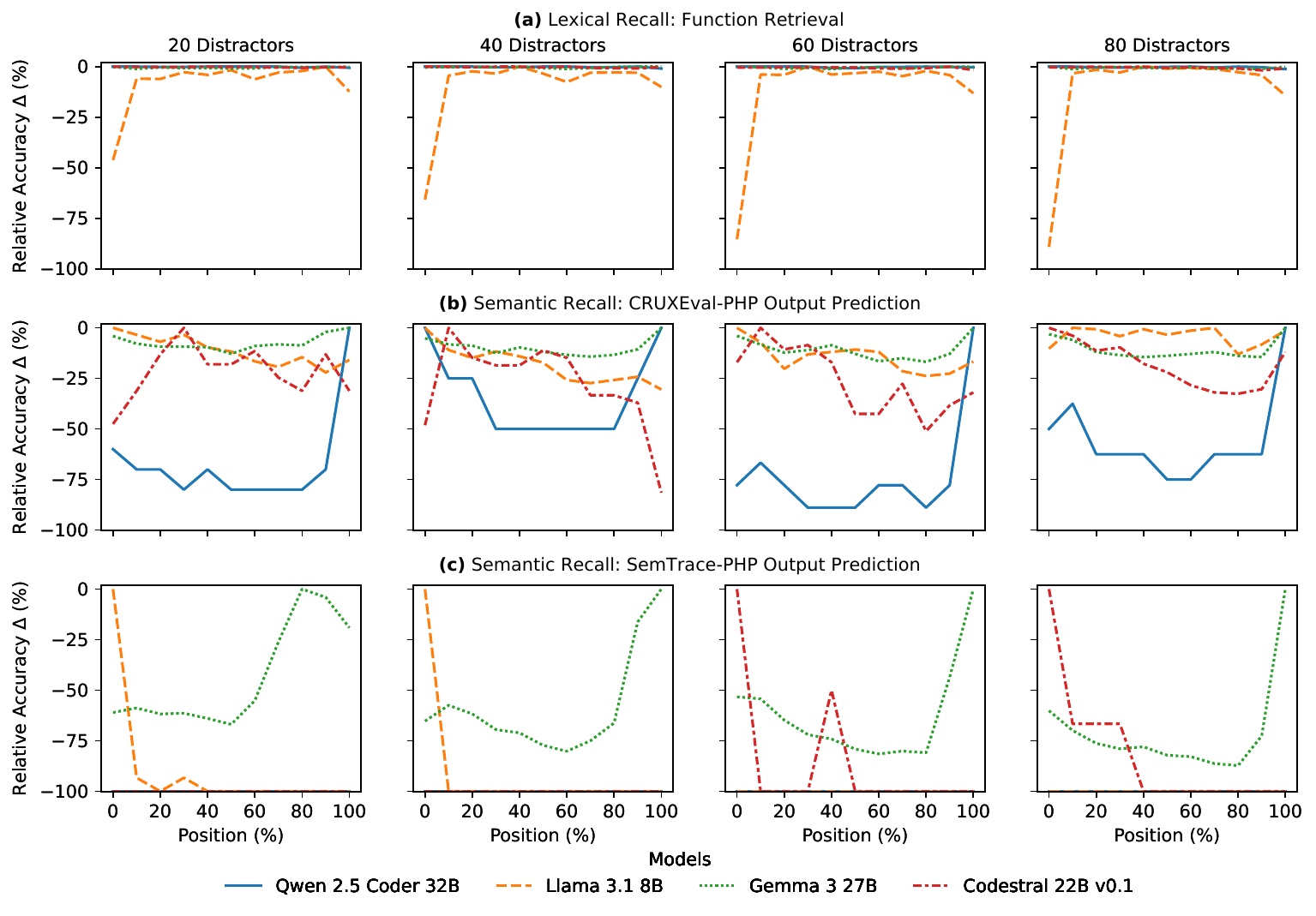}
    \caption{Dissociation between lexical and semantic recall across code positions in long contexts for PHP. The same fundamental pattern holds as for Python and JavaScript. Notably, on \semtracephp (Panel c), several models achieve near-zero baseline accuracy across all positions, despite performing better on \cruxevalphp than \cruxevaljs. We plot these cases at $-100$\% as a visualization convention; the relative drop is otherwise undefined because baseline accuracy is zero.}
    \label{fig: long context panel php}
\end{figure*}

\autoref{fig: long context panel js} and \autoref{fig: long context panel php} present the results of our experiments on \cruxevaljs/\semtracejs and \cruxevalphp/\semtracephp, respectively, using four representative models and the same experimental protocol as \cruxeval/\semtrace (detailed in~\autoref{sec: experimental setup}). Both languages exhibit the same fundamental dissociation as Python: lexical recall remains near-perfect and position-independent, while semantic recall degrades as target code moves toward the center of the context, with more severe degradation on \semtrace than on \cruxeval. The one notable exception is \semtracephp, where several models achieve near-zero accuracy regardless of position, which is surprising given their median performance on \cruxevalphp is better than \cruxevaljs. This might suggest that the degree of reliance of models on pattern-matching instead of semantic recall can also be impacted by the programming language in question. Despite this, the core finding that semantic recall is substantially more position-sensitive than lexical recall holds consistently across all three languages.

\section{Model and Dataset Licenses \label{app: licenses}}
We include the licenses for models and datasets in~\autoref{tab: model licenses} and~\autoref{tab: dataset licenses}, respectively.
\begin{table}
\centering
\begin{tabular}{ll}
\toprule
\textbf{Model Name} & \textbf{License}\\
\midrule
\codestral & MNLP-0.1\\
\gpt & Proprietary\\
\deepseeklarge & DeepSeek license\\
\deepseeksmall & DeepSeek license\\
\llamasmall & Llama 3.1 license\\
\llamalarge & Llama 3.3 license\\
\qwenlarge & Apache 2.0\\
\qwensmall & Apache 2.0\\
\gemmasmall & Gemma ToS\\
\gemmalarge & Gemma ToS\\
\bottomrule
\end{tabular}
\caption{\label{tab: model licenses} Model licensing information.}
\end{table}
\begin{table}
\centering
\begin{tabular}{ll}
\toprule
\textbf{Dataset Name} & \textbf{License}\\
\midrule
\cruxeval & MIT License \\
\codesearchnet & MIT License \\
\bottomrule
\end{tabular}
\caption{\label{tab: dataset licenses} Dataset licensing information.}
\end{table}

\section{Prompt Templates \label{app: prompt templates}}
In this section we include templates for input prediction (\autoref{fig: input prediction prompt template}), output prediction (\autoref{fig: output prediction prompt template}), and function-level retrieval (\autoref{fig: function retrieval prompt template}) tasks. We leverage assistant prefill such that each model provides a predictable and easy-to-parse response.
\begin{figure}
    \begin{tcolorbox}[
        width=\columnwidth,
        colback=white,
        colframe=black!20,
        boxrule=0.5pt,
        arc=1mm]
        \begin{small}
        \texttt{User:}\\
        \> \texttt{You are given a number of Python functions and an assertion containing an output of one of the functions. Find any input such that executing that function on the input leads to the given output. There may be multiple answers, but you should only output one.} \\
        \\
        \> \texttt{[ASSERTION]} \\
        \> \texttt{assert \{output\} == f(??)} \\
        \> \texttt{[/ASSERTION]} \\
        \\
        \> \texttt{[FUNCTIONS]} \\
        \> \texttt{\{code\_block\}} \\
        \> \texttt{[/FUNCTIONS]} \\
        \\
        \> \texttt{[ASSERTION]} \\
        \> \texttt{assert \{output\} == f(??)} \\
        \> \texttt{[/ASSERTION]} \\
        \\
        \> \texttt{You are given a number of Python functions and an assertion containing an output of one of the functions. Find any input such that executing that function on the input leads to the given output. There may be multiple answers, but you should only output one.} \\
        \\
        \texttt{Assistant:} \\
        \> \texttt{Sure! Here is the corresponding input:} \\
        \\
        \> \texttt{\mdtick\mdtick\mdtick python} \\
        \> \texttt{assert \{output\} == f(}
        \end{small}
    \end{tcolorbox}
    \caption{Input Prediction Chat Completion Prompt Template.}
    \label{fig: input prediction prompt template}
\end{figure}
\begin{figure}
    \begin{tcolorbox}[
        width=\columnwidth,
        colback=white,
        colframe=black!20,
        boxrule=0.5pt,
        arc=1mm]
        \begin{small}
        \texttt{User:}\\
        \> \texttt{You are given a number of Python functions and an assertion containing an input to one of the functions. Complete the assertion with a literal (no unsimplified expressions, no function calls) containing the output when executing the provided code on the given input, even if the function is incorrect or incomplete. Do NOT output any extra information.} \\
        \\
        \> \texttt{[ASSERTION]} \\
        \> \texttt{assert f(\{input\}) == ??} \\
        \> \texttt{[/ASSERTION]} \\
        \\
        \> \texttt{[FUNCTIONS]} \\
        \> \texttt{\{code\_block\}} \\
        \> \texttt{[/FUNCTIONS]} \\
        \\
        \> \texttt{[ASSERTION]} \\
        \> \texttt{assert f(\{input\}) == ??} \\
        \> \texttt{[/ASSERTION]} \\
        \\
        \> \texttt{You are given a number of Python functions and an assertion containing an input to one of the functions. Complete the assertion with a literal (no unsimplified expressions, no function calls) containing the output when executing the provided code on the given input, even if the function is incorrect or incomplete. Do NOT output any extra information.} \\
        \\
        \texttt{Assistant:} \\
        \> \texttt{Sure! Here is the corresponding output:} \\
        \\
        \> \texttt{\mdtick \mdtick \mdtick python} \\
        \> \texttt{assert f(\{input\}) == }
        \end{small}
    \end{tcolorbox}
    \caption{Output Prediction Chat Completion Prompt Template.}
    \label{fig: output prediction prompt template}
\end{figure}
\begin{figure}
    \begin{tcolorbox}[
        width=\columnwidth,
        colback=white,
        colframe=black!20,
        boxrule=0.5pt,
        arc=1mm]
        \begin{small}
        \texttt{User:}\\
        \> \texttt{Each line in the code block below starts with a random key. I'm looking for a function starting at key `\{start\}` and ending at key `\{end\}` in the code snippet below. Can you help me find it?} \\
        \\
        \> \texttt{\mdtick\mdtick\mdtick python} \\
        \> \texttt{\{code\_block\}} \\
        \> \texttt{\mdtick\mdtick\mdtick} \\
        \\
        \> \texttt{Each line in the code block above starts with a random key. I'm looking for a function starting at key `\{start\}` and ending at key `\{end\}` in the code snippet above. Can you help me find it?} \\
        \\
        \texttt{Assistant:} \\
        \> \texttt{Sure! Here is the full function starting at key `\{start\}` and ending at key `\{end\}`:} \\
        \\
        \> \texttt{\mdtick\mdtick\mdtick python}
        \end{small}
    \end{tcolorbox}
    \caption{Function-level Retrieval Chat Completion Prompt Template.}
    \label{fig: function retrieval prompt template}
\end{figure}

\section{Detailed Setup Information \label{app: detailed setup}}
\noindent{\textbf{Machine Details.}} We performed all experiments using an AWS EC2 \texttt{p5e.48xlarge} instance equipped with 192 cores, 2048GB RAM, and eight NVIDIA H200 GPUs on Ubuntu 22.04 with CUDA 12.4. To serve open-weight LLMs, we use vLLM 0.7.1~\cite{kwon2023efficient}. We access \gpt through the OpenAI API.

\noindent{\textbf{Execution Time.}}
For the final evaluation runs, we spent a total of 47 GPU-days on open-weight model inference. We spent about 2 days running experiments on \gpt through the OpenAI API. We estimate that total usage, including reruns and development, might be 1.5--2 times higher than our evaluation runs.

\end{document}